\pdfoutput=1
\documentclass[aps,pre,groupedaddress,nofootinbib,twocolumn,showkeys]{revtex4-2}

\usepackage[dvipsnames]{xcolor}
\usepackage{amsmath,amssymb}
\usepackage{graphicx}
\usepackage{footmisc}
\usepackage{bbold,soul}
\usepackage{lipsum}
\usepackage{enumitem}
\usepackage{hyperref}

\makeatletter
\newcommand\footnoteref[1]{\protected@xdef\@thefnmark{\ref{#1}}\@footnotemark}
\makeatother

\definecolor{darkblue}{rgb}{0.0,0,0.75}
\hypersetup{
   colorlinks=true,
   linkcolor=red,
   filecolor=black,      
   urlcolor=darkblue,
   citecolor=red,
}

\begin{document}

\title{When not to use machine learning: a perspective on potential and limitations}

\author{Matthew R. Carbone} \email{mcarbone@bnl.gov}
\affiliation{Computational Science Initiative, Brookhaven National Laboratory, Upton, New York 11973}

\date{\today}

\keywords{artificial intelligence; computation/computing; machine learning; modeling}

\begin{abstract}
The unparalleled success of artificial intelligence (AI) in the technology sector has catalyzed an enormous amount of research in the scientific community. It has proven to be a powerful tool, but as with any rapidly developing field, the deluge of information can be overwhelming, confusing and sometimes misleading. This can make it easy to become lost in the same hype cycles that have historically ended in the periods of scarce funding and depleted expectations known as AI Winters. Furthermore, while the importance of innovative, high-risk research cannot be overstated, it is also imperative to understand the fundamental limits of available techniques, especially in young fields where the rules appear to be constantly rewritten and as the likelihood of application to high-stakes scenarios increases. In this perspective, we highlight the guiding principles of data-driven modeling, how these principles imbue models with almost magical predictive power, and how they also impose limitations on the scope of problems they can address. Particularly, understanding when \textit{not} to use data-driven techniques, such as machine learning, is not something commonly explored, but is just as important as knowing how to apply the techniques properly. We hope that the discussion to follow provides researchers throughout the sciences with a better understanding of when said techniques are appropriate, the pitfalls to watch for, and most importantly, the confidence to leverage the power they can provide.
\end{abstract}

\maketitle

\section{Introduction} The story of artificial intelligence (AI) began in the mid 1900's, when computer scientists started considering a simple question: ``can machines think?"~\cite{Turingmachinery1950computing,jordan2015machine}. Ever since, the mythical goal of achieving true ``human-like" AI has framed decades of scientific conversation and has motivated many key breakthroughs, including the backbone of modern machine learning (ML) algorithms: neural networks~\cite{mcculloch1943logical,sarle1994,paliwal2009neural}. As the computational machinery required to realize the practical applications of ML would come decades later~\cite{bishop2006pattern}, its full potential would not yet be immediately understood. That time has come, and despite the turbulence of multiple ``AI winters" over the past decades~\cite{crevier1993ai,hendler2008avoiding,reyes2019machine}, we are currently living the AI/ML revolution.

Broadly, AI and ML are two related families of methods that fall under the larger ``data-driven" umbrella. Built upon well established theory in the statistics and applied mathematics communities~\cite{sarle1994}, modern-day AI and ML is best understood as the intersection of powerful modeling paradigms with ``big-data" and bleeding edge hardware (e.g. graphics processing units; GPUs). The general interpretation (though not the only one~\cite{langley2011changing}) is that AI is a superset of ML~\cite{holzinger2018current} and consists of techniques that are used to mimic human cognition and decision-making, whereas ML is more focused on the mathematical and numerical approaches. Often, ML is described as the ability of a program to learn a task without being programmed with task-specific heuristics~\cite{mitchell2014machine}. However, the distinction between AI and ML is not germane to most applications (and many applications use parts of both), hence the blanket term ``AI/ML" is used commonly as a substitute for ``data-driven" in many contexts. In this perspective, we will focus primarily on supervised ML, though many of the key points to come apply to data-driven approaches in general.

Cruising behind the slipstream created by tremendous success in the technology sector, ML has found wide applicability in the materials, chemical and physical sciences. For example, the discovery, characterization and design of new materials, molecules and nanoparticles~\cite{juhas2015complex,timoshenko2017supervised,butler2018machine,sanchez2018inverse,coley2018machine,gomez2018automatic,carbone2019classification,liu2019using,zhang2019unsupervised,torrisi2020random,mercado2021graph,mouchlis2021advances}, surrogate models for spectroscopy and other properties~\cite{gilmer2017neural,xie2018crystal,carbone2020machine,rankine2022accurate}, self-driving laboratories/autonomous experimentation~\cite{noack2020advances,macleod2020self,epps2020artificial,noack2021gaussian,bateni2022autonomous,D2DD00014H}, and neural network potentials~\cite{behler2007generalized,behler2011atom,behler2011neural,artrith2016implementation,behler2021four} have all been powered by ML and related methods. The current state of ML in materials science specifically has also been thoroughly documented in many excellent reviews~\cite{butler2018machine,gomes2019artificial,wei2019machine,morgan2020opportunities} that cover subject matter ranging from applications to computational screening and interpretation. On a related note, for technical details and timely tutorials, we refer those interested readers to Refs.~\onlinecite{wang2020machine} and \onlinecite{artrith2021best}. However, while the scope of ML-relevant problems is huge, not every problem can effectively leverage the power ML provides. Worse still, sometimes ML may seem to be a perfectly reasonable choice only to fail dramatically~\cite{sambasivan2021everyone}; such failures can often be traced back to the foundations of any ML tool: the data.

In this perspective, we ask and answer a foundational question which ultimately has \textit{everything} to do with data: when should you not use machine learning? ML is the jackhammer of the applied math world, and is able to channel incredible power provided by the interplay of highly flexible models, large databases and GPU-enabled supercomputers. But you wouldn't use a jackhammer to do brain surgery. At least for the time being, there are classes of problems for which ML is not well-suited~\cite{wang2020machine,chollet2021deep}. We address this issue not to dissuade researchers from using these methods, but rather to \textit{empower} them to do so correctly, and to avoid wasting valuable time and resources. Understanding the limitations and application spaces of our tools will help us build better ones, and solve larger problems more confidently and with more consistency.

\section{The devil's in the distance} Newcomers to the field of ML will find themselves immediately buried under an avalanche of enticing algorithms applicable to their scientific problem~\cite{ayodele2010types}. Many of these choices are so sophisticated that it is unreasonable to expect any ML non-expert to understand their finer nuances and how/why they can fail. The steep learning curve combined with their intrinsic complexity, mythical ``black box" nature and stunning ability to make accurate predictions can make ML appear almost magical. It may come as a surprise then that in spite of said complexities, almost all supervised ML models are paradigmatically the same and are built upon a familiar quantity: distance.

The supervised ML problem is one of minimizing the distance between predicted and true values mapped by an approximate function on the appropriate inputs. A distance can be a proper metric, such as the Euclidean or Manhattan norms, or something less pedestrian, such as a divergence (a distance measure between two distributions) or a cross-entropy loss function. Regardless, the principle is the same: consider un-regularized, supervised ML, where given a source of ground truth $F$ and a ML model $f_\theta,$ the goal is to find parameters $\theta$ such that the distance between $F(x)$ and $f_\theta(x)$ is as small as possible for all possible $x$ in some use case. While this is only one type of ML, most techniques share this common theme. For example, Deep Q reinforcement learning~\cite{mnih2015human} leverages neural networks to map states (inputs) to decisions (outputs), and unsupervised learning algorithms rely on the same notion of distance to perform clustering and dimensionality reduction that supervised learning techniques use to minimize loss functions. Variational autoencoders~\cite{kingma_auto-encoding_2014,higgins2016beta,miles2021machine} try not only to minimize reconstruction loss, but simultaneously keep a compressed, latent representation as close to some target distribution as possible (usually for use in generative applications). Numerical optimization is the engine that systematically tunes model parameters $\theta$ in gradient-based ML,\footnote{The classic numerical optimizer is gradient/stochastic gradient descent, with more recently established developments showing systematic improvements in deep learning, e.g. Adam~\cite{kingma2014adam}.} and its \textit{only} objective is to minimize some measure of distance between ground truth and model predictions. 

Additionally, in order to increase the confidence that ML models will be successful for a given task, it helps if the desired function is smooth: i.e. a small change in a feature should ideally correspond to a relatively small change in the target. This idea is more readily defined for regression than for classification, and the data being amenable to gradient-based methods is is not strictly required for ML to be successful. For example, e.g. Random Forests are not generally trained using gradient-based optimizers, but satisfying this requirement will usually help models generalize more effectively. The distance between the features of any two points of data is informed entirely by their numerical vector representation, and while these representations can be intuitive or human-interpretable, they must be mathematically rigorous.

The devil here, so to speak, is that what might appear intuitive to the experimenter may not be to the machine. For example, consider the problem of discovering novel molecules with certain properties. Molecules can be first encoded in string format (e.g. SMILES~\cite{weininger1988smiles}), and then a numerical latent representation. The structure of this latent space is informed by some target property~\cite{sanchez2018inverse}, and because any point in the latent space is just a numeric vector living in a vector space, a distance can be easily defined. This powerful encoding method can be used to ``interpolate between molecules" and thus discover new ones that perhaps we haven't previously considered, but it still relies on the principle of distance, both between molecules in the compressed latent space, and their target properties.

\begin{figure}[tb]
    \centering
    \includegraphics[width=\columnwidth]{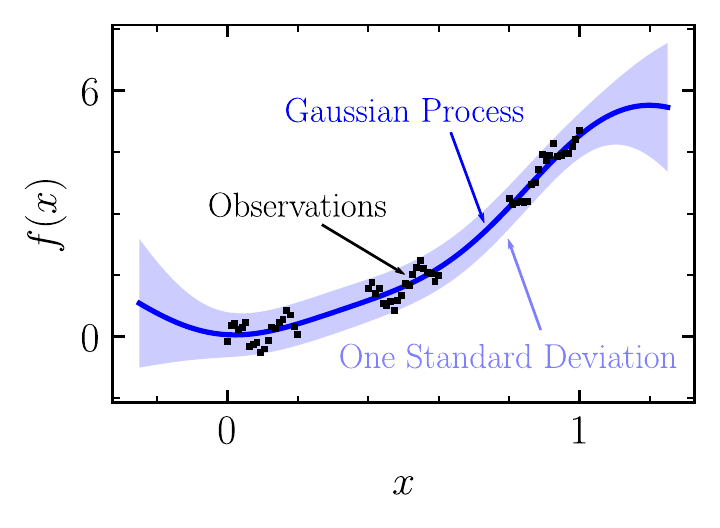}
    \caption{\label{fig:distance} A Gaussian Process, with a radial basis function kernel, fit to example data $F(x) = 5x^2 + \sin(50x)/3 + \mathcal{N}(\mu=0, \sigma=0.1).$ The small, noisy, high-frequency oscillations are not well-fit by a Gaussian Process, as the primary length scale in the data is defined by the distances between the three clusters, not in between them. Sophisticated, non-isotropic kernels could model both trends in principle, but the construction of such kernels often requires significant prior information about the problem, the knowledge of which might make modeling unnecessary.}
\end{figure}

Concretely, the length scales for differentiating between data points in the feature space are set by the corresponding targets. Large changes in target values between data points can cause ML models to ``focus" on the changes in the input space that caused it, possibly at the expense of failing to capture small changes. This is often referred to as the bias-variance trade-off. Most readers may be familiar with the concept of over-fitting: for instance, essentially any set of observations can be fit exactly by an arbitrarily high-order polynomial, but doing so will produce wildly varying results during inference and be unlikely to have captured anything meaningful about the underlying function. Conversely, a linear fit will only capture the most simple trends to the point of being useless for any nonlinear phenomena. Fig.~\ref{fig:distance} showcases a common middle ground, where the primary trend of the data is captured by a Gaussian Process~\cite{Rasmussen2006GP}, and smaller fluctuations are understood as noisy variations around that trend.

Consider a more realistic example: the Materials Project~\cite{Jain2013} database contains many geometry-relaxed structures, each with different compositions, space groups and local symmetries at 0 Kelvin. Thus, within this database, changes in e.g. the optical properties of these materials is primarily due to the aforementioned structural differences and not due to thermal disorder (i.e. distortions) one would find when running a molecular dynamics simulation. A ML model trained on this data could be limited in the sense that changes in certain structural motifs would be well-captured and others would not, necessitating caution when scoping its effective use cases. Conversely, in data-driven modeling, considering the distances between data points in both the input and output spaces, and thus which changes in your features are most contributing to the variance in the targets, can be instrumental in constructing effective training sets targeted to specific applications.

Inverse problems, and cases with extremely low signal-to-noise ratios (SNR), are another case in which the usefulness of ML varies significantly problem-to-problem. ML can only model functions: cases in which each input maps to a unique output. Inverse problems, or those in which a signal is used to resolve its source, are often ill-posed, making it challenging if not impossible to represent them by functions. In these cases, ML is not immediately appropriate, and the problem statement must be refined until a ``non-degenerate" subspace is found and can be modeled by a function. Once this space is identified, ML can excel, because it can pick out subtle patterns in this space where heuristics or human intuition may fail to do so, but the developed mapping \textit{must} be a function. This can also be understood in the language of distance: in the inverse problem, an ``epsilon-small" change in the input can result in an extremely large change in the output.\footnote{In the inverse problem, an ``input" might be an observable, such as a spectrum, and the output, a structure, such as in the structure refinement problem.} For example, consider phase retrieval in coherent diffraction imaging~\cite{zhang2016phase}. If a detector measures only the intensity of the signal, all phase information is lost by definition, and unless correlations between the intensity and phase exist (which are specific to the sub-problem of interest) and permit such an inverse mapping~\cite{wu2021three}, there is no way to confidently retrieve the phase information. The degree to which this is possible in general depends entirely on the specific system and the data available.

The same issue can be found when the SNR is low or close to 1: no data-driven technique will be useful if targets cannot be distinguished from each other. Relatedly, if the uncertainty during inference on an inverse problem can be \textit{accurately} quantified, then it is possible to use uncertainty-aware models to make predictions with error estimations~\cite{srivastava2014dropout,wilson2020bayesian,jospin2022hands}. However, in cases where the inversion is sufficiently ill-posed, the SNR will be so low that different results cannot be resolved. So while it might be possible to make predictions with error bars, they may not be useful.

In summary, it is always an instructive exercise to consider distances between the properties of entries in your database when attempting any data-driven modeling. The key is to fully understand the property of interest, and to choose a database and encoding such that different data points are sufficiently discriminatory with respect to the target property. If this is not possible, no data-driven method will be able to perform effectively. In these cases, utilization of other prior knowledge, more effective data screening or simply another type of technique entirely might be required. We note that there is no harm in simply trying new techniques on e.g. inverse problems where it is unclear how much pertinent information is present in a database that could allow an inverse mapping to be learned. Such possibility, however, should never be confused with certainty.

\section{On the importance of data distributions} To quantify the effectiveness of trained models, evaluation should always conclude with the presentation of evaluation metrics collected on a ``testing" subset of the database. In order to avoid training and hyperparameter tuning\footnote{Hyperparameters are un-trained parameters of the model and training procedure, such as the choice of loss function, or the number of neurons in some layer of a neural network.} biases, the testing set should be disjoint from the set of data that was used to train and tune the models (the training and cross-validation databases, respectively). Put even more simply, the rule of thumb is to ``blind" yourself from bias as best you can: take a chunk of data from the full space of data of interest, and not use it, look at it, or otherwise glean any information from it until you are ready to present results on a trained model and completed pipeline. As long as information from the testing set is not used during development, any approach to model tuning is appropriate (though some, such as using cross-validation, are highly recommended). These are the best practices which are critical to any successful ML project, and they are often highlighted in more technical tutorials~\cite{wang2020machine,artrith2021best} (and in more technical detail than presented here), but this is not the complete story.

The testing set is almost always discussed as an unbiased sample, a litmus test for how the model will perform on data it hasn't seen before. However, there is another use for the testing set: it should represent the real-world deployment scenario. In other words, the testing set should not only be disjoint from data the model and pipeline have seen before, it should also ideally represent the data on which the model must be performant. It is paramount to keep in mind that these two uses of the testing set are not always the same.

\begin{figure}[tb]
    \centering
    \includegraphics[width=\columnwidth]{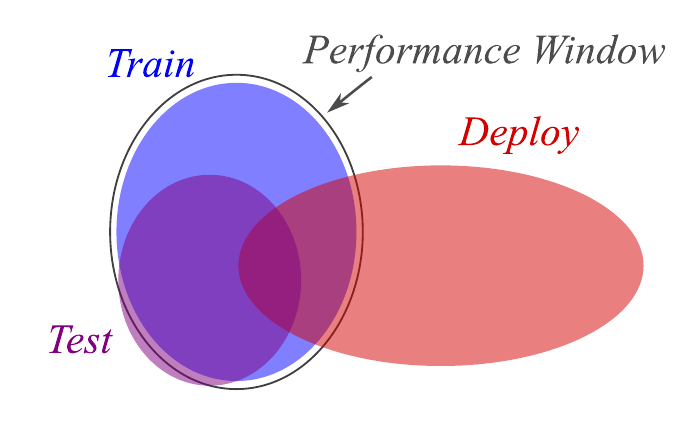}
    \caption{\label{fig:ven-diagram} A possible failure scenario in data-driven modeling: the model is fit to the training (including cross-validation) data, and evaluated on a testing set. The testing set is sampled (but is disjoint) from the same data as the training set, resulting in strong distribution-wise overlap with the training data and likely good performance. The deployment case is sampled from a different database, the distribution of which may only partially overlap with data the model is familiar with.}
\end{figure}

If the training data comes from a different distribution than data from your deployment scenario, it is highly likely the trained model will fail. Human intuition can actually take us far here, as there is a way to easily sanity-check if any two sets of data do \textit{not} come from the same distribution. Simply re-combine them and sample randomly. If you can easily tell the difference (i.e., determine from which distribution a sample originated), your testing set is ``out-of-sample" with respect to the training set. This will not always be the case (see e.g. adversarial examples, where human-imperceptible modifications to images can cause otherwise highly accurate ML models to go haywire~\cite{goodfellow2014explaining}), but in many scientific problems, it is a critical exercise to perform when planning a research campaign. For example, this can often happen when attempting to train a model on computer-simulated data (which is relatively cheap to obtain) and then deploying it on experimental data (which often requires expensive and time-consuming experiments). Such a use case is common in science, since we tend to have much less data available than in e.g. image recognition problems in computer science (where more labeled data can be simply bought). Indeed, there are some cases where the simulation is sufficiently accurate when compared to experimental measurements (e.g. predicting the space group from pair distribution data~\cite{liu2019using} or nanoparticle sizes from x-ray absorption spectra; XAS~\cite{timoshenko2017supervised}). Other cases will not work nearly as well, such as when comparing experimental and simulated XAS across a diverse crystal structure database~\cite{carbone2019classification}. Fig.~\ref{fig:ven-diagram} showcases this possible failure scenario. 

While it is not a hard-and-fast rule, one should never take for granted that data-driven models fit on one set of data will perform well on another. On the contrary, ML performs best (and is most powerful) when it is explicitly fit on the same type of data it is expected to perform on. This is a limitation that is often interpreted as weakness; on the contrary, this is actually a strength. For example, one high-impact example demonstrating this feature is that of neural network potentials~\cite{behler2007generalized,behler2011atom,behler2011neural,artrith2016implementation,behler2021four}, where the space of possible atomic configurations is kept small (potentials are fit on specific systems), and configurations are revisited throughout long molecular dynamics simulations. Potentials fit on one system are not expected to perform well on another, but they do perform to desired accuracy on the systems they're trained on.

Ensuring proper data selection is not a technical challenge, it is a human one~\cite{sambasivan2021everyone}. When considering if ML is the right tool for your problem the real-world deployment scenario must be considered. When at all possible, one should simply fit the model on data from the same distribution as in said deployment. In cases where there is not enough deployment data to fit models on (or to do transfer learning~\cite{weiss2016survey}) it is unlikely ML methods will perform well. If the scenario outlined in Fig.~\ref{fig:ven-diagram} is a possibility, sufficient labeled data from the deployment case \textit{must} be available to validate that the model is working properly. If these criteria cannot be satisfied, then there is no feasible way to validate the trained models in the desired use cases. Consequently, if the model's performance cannot be verified, it cannot be used.

\section{The data-driven ``No Free Lunch Theorem"} The considerable flexibility and information capacity of modern ML models (such as neural networks) comes at a cost. While exceptional at ``interpolating" within and close to the ``convex hull" defined by the boundaries of the training set, they will often fail in spectacular fashion when tasked with predicting outside of this region~\cite{behler2011atom,chollet2021deep}. This limitation can be partially addressed by the encoding of prior belief, which can take many forms, such as the functional form of the model, correlation information between input features, or boundary behavior. The information content of prior belief can be a game-changer: indeed, even the term ``data-driven" can be somewhat misleading~\cite{reyes2019machine} (though the description ``information-driven", while perhaps more accurate, may be a bit too ambiguous). That said, no data-driven model can make reliable predictions outside of the union of the data and prior information it was trained on. This information-theoretic perspective is tautological, but is often overlooked despite its significant implications: data-driven models generalize, they do not extrapolate beyond the aforementioned union with any reliability.

It is important to keep in mind that there is a difference between some trained ML model and an overhead algorithm that is operating for a particular use case which might be using one or more trained models. Often, these algorithms will involve a re-training step in which the inference or decision-making model is continuously updated. For example, reinforcement learning involves an ``outer loop" in which the environment is probed and feedback acquired through a reward function, decisions are made, and the decision maker refined. Gaussian Processes and ensemble methods are excellent choices for sampling new data because they naturally quantify uncertainty. Data can be sampled where uncertainty, and thus the likelihood of out-of-sample data, is high. Computer scientists already have a name for this: active learning. This can help the user understand where the model is predicting outside of it's information-theoretic interpolation window, and shore up the model's weaknesses by adding new data to the training set. Another way to interpret active learning is that the algorithm itself is detecting when it is extrapolating, and actively expanding its interpolation window to compensate.

\begin{figure}[tb]
    \centering
    \includegraphics[width=\columnwidth]{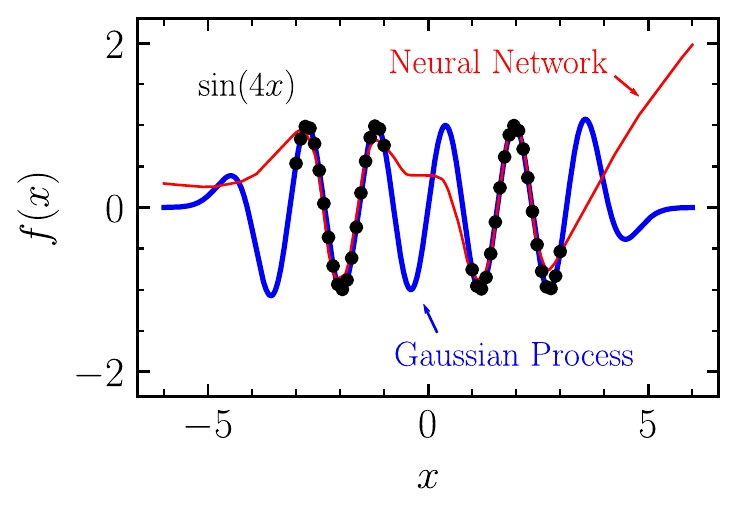}
    \caption{\label{fig:sine} A simple example of fitting example data (black) with two models: a simple neural network (red) and Gaussian Process with a periodic kernel (blue).}
    \label{fig:pyramid}
\end{figure}

To demonstrate the point, consider the apparently simple 1-dimensional example in Fig.~\ref{fig:sine}, where a neural network and Gaussian Process are tasked with modeling the function $F(x) = \sin(4x)$ with minimal observations. The neural network was trained only to minimize the mean squared error loss function on the provided data. Thus, it ends up fitting the data quite well around where data exists, but fails completely when far away from those points. These failures are essentially random and unpredictable, and are due to the particular state of the neural network's weights. Even ``within" the convex hull of the training data, in the region $x \in [-1, 1],$ the neural network does not perform, because the optimizer has no incentive to focus on that region (due to lack of ground truth information). On the other hand, the Gaussian Process is trained using kernels that explicitly encode correlation lengths, limiting the possible set of interpolating functions in the region $x \in [-1, 1].$ This ultimately results in a much better fit, consistent with the learned length scale of the kernel and the data used to fit it. That said, any model that operates on the Bayesian paradigm of starting with a prior belief which is then updated to a posterior when fit on data will revert back to the mean when sufficiently outside of the space of the data it is fit on; this is clearly observed here (the mean of the prior is 0).

One other important observation is that despite the natural human conclusion that the data is likely periodic (which we reach by simply looking at the black markers in Fig.~\ref{fig:sine}), the neural network cannot intuit this without prior knowledge. Thousands more data points could be provided, spanning a much larger range in $x,$ but no amount of data will encode periodicity in a way that a data-driven model understands. On the other hand, when encoded as a prior belief, either in the form of a kernel or perhaps even through explicit selection of a periodic function, only a handful of data points are necessary to completely characterize it. Expecting the model to understand periodicity without explicitly imbuing it with said information is akin to information-theoretic extrapolation: a non-starter.

Unfortunately, most modeling problems of interest are not quite as simple as the 1-dimensional example in Fig.~\ref{fig:sine}. Often, they have much higher-dimensional inputs and outputs, and display complicated non-linear behavior. It is much harder to visualize and interpret results in these situations (though tools are available, such as dimensionality reduction, e.g. Principal Component Analysis), and intuition derived from trial and error is usually the go-to method for understanding when and why the model is not performing well. At the least, care should be taken when formulating data-driven solutions to ensure that the information used to fit the models is carefully thought through. If it is possible the deployment scenario will include out-of-sample data, uncertainty quantification and active learning should be considered. Most of all, the way that humans understand and process data is very different from the way ML algorithms do, and that should never be overlooked.

\section{Outlook: avoiding the next AI Winter} AI/ML has been transformative in our society over the past decade. Especially in the technology sector, it has been applied with unnervingly surgical accuracy in targeted advertising, image recognition and neural translation, just to name a few examples. In these problem spaces, AI/ML is massively successful because they leverage its greatest strengths: the ability to process huge databases, and complex pattern recognition. Naturally, the scientific community has taken note and applied AI/ML to great effect in research acceleration and discovery. However, these techniques are not a cure-all, and cannot be applied to every problem. Without a doubt, we should keep pushing the boundaries of how we can apply AI/ML in science, but expectations should be kept appropriately measured.

The original AI Winters were caused by outrageously inflated expectations, spurred on by the promise of true AI, and while the actual winter always returns, it is hard to say if another \textit{AI} Winter is on the horizon~\cite{hendler2008avoiding,floridi2020ai}. In retrospect, the original idea of creating a synthetic autonomous thinker akin to a human was incredibly arrogant. After all, we're competing with millions of years of evolutionary instinct and development, including the most complex black box AI/ML algorithm we know of: the human brain. We would pose a simple question: why compete when we can collaborate?

One day, humanity will likely create sentient AI,\footnote{For the sake of argument, we will take humanity's survival for granted.} but we're not there yet, and for better or for worse, we're not even close. What we \textit{do} have is a wonderful suite of data-driven tools, including those found in the domain of AI/ML, which have the potential to significantly accelerate scientific research and discovery. These tools are meant to empower the experimenter, not to replace them. For the foreseeable future, we must still rely on human researchers to start problems with scientific hypotheses, find appropriate use cases for data-driven tools, and to apply them properly. AI/ML is not magic, and it is of the utmost importance, not only for each individual research project but also to the future of AI/ML in science, that its potential is never taken for granted.

\section*{Acknowledgements} MRC would like to thank Steven B. Torrisi, Marco Baity-Jesi, Phillip M. Maffettone and Chuntian Cao, for their critical feedback during the writing of this perspective. This material is based upon work supported by the U.S. Department of Energy, Office of Science, Office of Basic Energy Sciences, under Award Numbers FWP PS-030 and DE-SC-0012704.

\section*{Conflicts of interest} MRC declares no conflicts of interest.


%


\begin{thebibliography}{63}%
\makeatletter
\providecommand \@ifxundefined [1]{%
 \@ifx{#1\undefined}
}%
\providecommand \@ifnum [1]{%
 \ifnum #1\expandafter \@firstoftwo
 \else \expandafter \@secondoftwo
 \fi
}%
\providecommand \@ifx [1]{%
 \ifx #1\expandafter \@firstoftwo
 \else \expandafter \@secondoftwo
 \fi
}%
\providecommand \natexlab [1]{#1}%
\providecommand \enquote  [1]{``#1''}%
\providecommand \bibnamefont  [1]{#1}%
\providecommand \bibfnamefont [1]{#1}%
\providecommand \citenamefont [1]{#1}%
\providecommand \href@noop [0]{\@secondoftwo}%
\providecommand \href [0]{\begingroup \@sanitize@url \@href}%
\providecommand \@href[1]{\@@startlink{#1}\@@href}%
\providecommand \@@href[1]{\endgroup#1\@@endlink}%
\providecommand \@sanitize@url [0]{\catcode `\\12\catcode `\$12\catcode
  `\&12\catcode `\#12\catcode `\^12\catcode `\_12\catcode `\%12\relax}%
\providecommand \@@startlink[1]{}%
\providecommand \@@endlink[0]{}%
\providecommand \url  [0]{\begingroup\@sanitize@url \@url }%
\providecommand \@url [1]{\endgroup\@href {#1}{\urlprefix }}%
\providecommand \urlprefix  [0]{URL }%
\providecommand \Eprint [0]{\href }%
\providecommand \doibase [0]{http://dx.doi.org/}%
\providecommand \selectlanguage [0]{\@gobble}%
\providecommand \bibinfo  [0]{\@secondoftwo}%
\providecommand \bibfield  [0]{\@secondoftwo}%
\providecommand \translation [1]{[#1]}%
\providecommand \BibitemOpen [0]{}%
\providecommand \bibitemStop [0]{}%
\providecommand \bibitemNoStop [0]{.\EOS\space}%
\providecommand \EOS [0]{\spacefactor3000\relax}%
\providecommand \BibitemShut  [1]{\csname bibitem#1\endcsname}%
\let\auto@bib@innerbib\@empty
\bibitem [{\citenamefont {Turing}(1950)}]{Turingmachinery1950computing}%
  \BibitemOpen
  \bibfield  {author} {\bibinfo {author} {\bibfnamefont {A.~M.}\ \bibnamefont
  {Turing}},\ }\href@noop {} {\bibfield  {journal} {\bibinfo  {journal} {Mind}\
  }\textbf {\bibinfo {volume} {59}},\ \bibinfo {pages} {433} (\bibinfo {year}
  {1950})}\BibitemShut {NoStop}%
\bibitem [{\citenamefont {Jordan}\ and\ \citenamefont
  {Mitchell}(2015)}]{jordan2015machine}%
  \BibitemOpen
  \bibfield  {author} {\bibinfo {author} {\bibfnamefont {M.~I.}\ \bibnamefont
  {Jordan}}\ and\ \bibinfo {author} {\bibfnamefont {T.~M.}\ \bibnamefont
  {Mitchell}},\ }\href@noop {} {\bibfield  {journal} {\bibinfo  {journal}
  {Science}\ }\textbf {\bibinfo {volume} {349}},\ \bibinfo {pages} {255}
  (\bibinfo {year} {2015})}\BibitemShut {NoStop}%
\bibitem [{\citenamefont {McCulloch}\ and\ \citenamefont
  {Pitts}(1943)}]{mcculloch1943logical}%
  \BibitemOpen
  \bibfield  {author} {\bibinfo {author} {\bibfnamefont {W.~S.}\ \bibnamefont
  {McCulloch}}\ and\ \bibinfo {author} {\bibfnamefont {W.}~\bibnamefont
  {Pitts}},\ }\href@noop {} {\bibfield  {journal} {\bibinfo  {journal} {Bull
  Math Biophys.}\ }\textbf {\bibinfo {volume} {5}},\ \bibinfo {pages} {115}
  (\bibinfo {year} {1943})}\BibitemShut {NoStop}%
\bibitem [{\citenamefont {Sarle}(1994)}]{sarle1994}%
  \BibitemOpen
  \bibfield  {author} {\bibinfo {author} {\bibfnamefont {W.~S.}\ \bibnamefont
  {Sarle}},\ }in\ \href@noop {} {\emph {\bibinfo {booktitle} {Proceedings of
  the Nineteenth Annual SAS Users Groups International Conference}}}\ (\bibinfo
  {organization} {SAS Institute, Inc.},\ \bibinfo {year} {1994})\ p.\ \bibinfo
  {pages} {1538–1550}\BibitemShut {NoStop}%
\bibitem [{\citenamefont {Paliwal}\ and\ \citenamefont
  {Kumar}(2009)}]{paliwal2009neural}%
  \BibitemOpen
  \bibfield  {author} {\bibinfo {author} {\bibfnamefont {M.}~\bibnamefont
  {Paliwal}}\ and\ \bibinfo {author} {\bibfnamefont {U.~A.}\ \bibnamefont
  {Kumar}},\ }\href@noop {} {\bibfield  {journal} {\bibinfo  {journal} {Expert
  Syst. Appl.}\ }\textbf {\bibinfo {volume} {36}},\ \bibinfo {pages} {2}
  (\bibinfo {year} {2009})}\BibitemShut {NoStop}%
\bibitem [{\citenamefont {Bishop}\ and\ \citenamefont
  {Nasrabadi}(2006)}]{bishop2006pattern}%
  \BibitemOpen
  \bibfield  {author} {\bibinfo {author} {\bibfnamefont {C.~M.}\ \bibnamefont
  {Bishop}}\ and\ \bibinfo {author} {\bibfnamefont {N.~M.}\ \bibnamefont
  {Nasrabadi}},\ }\href@noop {} {\emph {\bibinfo {title} {Pattern recognition
  and machine learning}}},\ Vol.~\bibinfo {volume} {4}\ (\bibinfo  {publisher}
  {Springer},\ \bibinfo {year} {2006})\BibitemShut {NoStop}%
\bibitem [{\citenamefont {Crevier}(1993)}]{crevier1993ai}%
  \BibitemOpen
  \bibfield  {author} {\bibinfo {author} {\bibfnamefont {D.}~\bibnamefont
  {Crevier}},\ }\href@noop {} {\emph {\bibinfo {title} {AI: the tumultuous
  history of the search for artificial intelligence}}}\ (\bibinfo  {publisher}
  {Basic Books, Inc.},\ \bibinfo {year} {1993})\BibitemShut {NoStop}%
\bibitem [{\citenamefont {Hendler}(2008)}]{hendler2008avoiding}%
  \BibitemOpen
  \bibfield  {author} {\bibinfo {author} {\bibfnamefont {J.}~\bibnamefont
  {Hendler}},\ }\href@noop {} {\bibfield  {journal} {\bibinfo  {journal} {IEEE
  Intell. Syst.}\ }\textbf {\bibinfo {volume} {23}},\ \bibinfo {pages} {2}
  (\bibinfo {year} {2008})}\BibitemShut {NoStop}%
\bibitem [{\citenamefont {Reyes}\ and\ \citenamefont
  {Maruyama}(2019)}]{reyes2019machine}%
  \BibitemOpen
  \bibfield  {author} {\bibinfo {author} {\bibfnamefont {K.~G.}\ \bibnamefont
  {Reyes}}\ and\ \bibinfo {author} {\bibfnamefont {B.}~\bibnamefont
  {Maruyama}},\ }\href@noop {} {\bibfield  {journal} {\bibinfo  {journal} {MRS
  Bull.}\ }\textbf {\bibinfo {volume} {44}},\ \bibinfo {pages} {530} (\bibinfo
  {year} {2019})}\BibitemShut {NoStop}%
\bibitem [{\citenamefont {Langley}\ \emph {et~al.}(2011)\citenamefont {Langley}
  \emph {et~al.}}]{langley2011changing}%
  \BibitemOpen
  \bibfield  {author} {\bibinfo {author} {\bibfnamefont {P.}~\bibnamefont
  {Langley}} \emph {et~al.},\ }\href@noop {} {\bibfield  {journal} {\bibinfo
  {journal} {Mach. Learn.}\ }\textbf {\bibinfo {volume} {82}},\ \bibinfo
  {pages} {275} (\bibinfo {year} {2011})}\BibitemShut {NoStop}%
\bibitem [{\citenamefont {Holzinger}\ \emph {et~al.}(2018)\citenamefont
  {Holzinger}, \citenamefont {Kieseberg}, \citenamefont {Weippl},\ and\
  \citenamefont {Tjoa}}]{holzinger2018current}%
  \BibitemOpen
  \bibfield  {author} {\bibinfo {author} {\bibfnamefont {A.}~\bibnamefont
  {Holzinger}}, \bibinfo {author} {\bibfnamefont {P.}~\bibnamefont
  {Kieseberg}}, \bibinfo {author} {\bibfnamefont {E.}~\bibnamefont {Weippl}}, \
  and\ \bibinfo {author} {\bibfnamefont {A.~M.}\ \bibnamefont {Tjoa}},\ }in\
  \href@noop {} {\emph {\bibinfo {booktitle} {International Cross-Domain
  Conference for Machine Learning and Knowledge Extraction}}}\ (\bibinfo
  {organization} {Springer},\ \bibinfo {year} {2018})\ pp.\ \bibinfo {pages}
  {1--8}\BibitemShut {NoStop}%
\bibitem [{\citenamefont {Mitchell}(2014)}]{mitchell2014machine}%
  \BibitemOpen
  \bibfield  {author} {\bibinfo {author} {\bibfnamefont {J.~B.}\ \bibnamefont
  {Mitchell}},\ }\href@noop {} {\bibfield  {journal} {\bibinfo  {journal}
  {Wiley Interdiscip. Rev. Comput. Mol. Sci.}\ }\textbf {\bibinfo {volume}
  {4}},\ \bibinfo {pages} {468} (\bibinfo {year} {2014})}\BibitemShut {NoStop}%
\bibitem [{\citenamefont {Juh{\'a}s}\ \emph {et~al.}(2015)\citenamefont
  {Juh{\'a}s}, \citenamefont {Farrow}, \citenamefont {Yang}, \citenamefont
  {Knox},\ and\ \citenamefont {Billinge}}]{juhas2015complex}%
  \BibitemOpen
  \bibfield  {author} {\bibinfo {author} {\bibfnamefont {P.}~\bibnamefont
  {Juh{\'a}s}}, \bibinfo {author} {\bibfnamefont {C.~L.}\ \bibnamefont
  {Farrow}}, \bibinfo {author} {\bibfnamefont {X.}~\bibnamefont {Yang}},
  \bibinfo {author} {\bibfnamefont {K.~R.}\ \bibnamefont {Knox}}, \ and\
  \bibinfo {author} {\bibfnamefont {S.~J.}\ \bibnamefont {Billinge}},\
  }\href@noop {} {\bibfield  {journal} {\bibinfo  {journal} {Acta Crystallogr.
  A}\ }\textbf {\bibinfo {volume} {71}},\ \bibinfo {pages} {562} (\bibinfo
  {year} {2015})}\BibitemShut {NoStop}%
\bibitem [{\citenamefont {Timoshenko}\ \emph {et~al.}(2017)\citenamefont
  {Timoshenko}, \citenamefont {Lu}, \citenamefont {Lin},\ and\ \citenamefont
  {Frenkel}}]{timoshenko2017supervised}%
  \BibitemOpen
  \bibfield  {author} {\bibinfo {author} {\bibfnamefont {J.}~\bibnamefont
  {Timoshenko}}, \bibinfo {author} {\bibfnamefont {D.}~\bibnamefont {Lu}},
  \bibinfo {author} {\bibfnamefont {Y.}~\bibnamefont {Lin}}, \ and\ \bibinfo
  {author} {\bibfnamefont {A.~I.}\ \bibnamefont {Frenkel}},\ }\href@noop {}
  {\bibfield  {journal} {\bibinfo  {journal} {J. Phys. Chem. Lett.}\ }\textbf
  {\bibinfo {volume} {8}},\ \bibinfo {pages} {5091} (\bibinfo {year}
  {2017})}\BibitemShut {NoStop}%
\bibitem [{\citenamefont {Butler}\ \emph {et~al.}(2018)\citenamefont {Butler},
  \citenamefont {Davies}, \citenamefont {Cartwright}, \citenamefont {Isayev},\
  and\ \citenamefont {Walsh}}]{butler2018machine}%
  \BibitemOpen
  \bibfield  {author} {\bibinfo {author} {\bibfnamefont {K.~T.}\ \bibnamefont
  {Butler}}, \bibinfo {author} {\bibfnamefont {D.~W.}\ \bibnamefont {Davies}},
  \bibinfo {author} {\bibfnamefont {H.}~\bibnamefont {Cartwright}}, \bibinfo
  {author} {\bibfnamefont {O.}~\bibnamefont {Isayev}}, \ and\ \bibinfo {author}
  {\bibfnamefont {A.}~\bibnamefont {Walsh}},\ }\href@noop {} {\bibfield
  {journal} {\bibinfo  {journal} {Nature}\ }\textbf {\bibinfo {volume} {559}},\
  \bibinfo {pages} {547} (\bibinfo {year} {2018})}\BibitemShut {NoStop}%
\bibitem [{\citenamefont {Sanchez-Lengeling}\ and\ \citenamefont
  {Aspuru-Guzik}(2018)}]{sanchez2018inverse}%
  \BibitemOpen
  \bibfield  {author} {\bibinfo {author} {\bibfnamefont {B.}~\bibnamefont
  {Sanchez-Lengeling}}\ and\ \bibinfo {author} {\bibfnamefont {A.}~\bibnamefont
  {Aspuru-Guzik}},\ }\href@noop {} {\bibfield  {journal} {\bibinfo  {journal}
  {Science}\ }\textbf {\bibinfo {volume} {361}},\ \bibinfo {pages} {360}
  (\bibinfo {year} {2018})}\BibitemShut {NoStop}%
\bibitem [{\citenamefont {Coley}\ \emph {et~al.}(2018)\citenamefont {Coley},
  \citenamefont {Green},\ and\ \citenamefont {Jensen}}]{coley2018machine}%
  \BibitemOpen
  \bibfield  {author} {\bibinfo {author} {\bibfnamefont {C.~W.}\ \bibnamefont
  {Coley}}, \bibinfo {author} {\bibfnamefont {W.~H.}\ \bibnamefont {Green}}, \
  and\ \bibinfo {author} {\bibfnamefont {K.~F.}\ \bibnamefont {Jensen}},\
  }\href@noop {} {\bibfield  {journal} {\bibinfo  {journal} {Acc. Chem. Res.}\
  }\textbf {\bibinfo {volume} {51}},\ \bibinfo {pages} {1281} (\bibinfo {year}
  {2018})}\BibitemShut {NoStop}%
\bibitem [{\citenamefont {G{\'o}mez-Bombarelli}\ \emph
  {et~al.}(2018)\citenamefont {G{\'o}mez-Bombarelli}, \citenamefont {Wei},
  \citenamefont {Duvenaud}, \citenamefont {Hern{\'a}ndez-Lobato}, \citenamefont
  {S{\'a}nchez-Lengeling}, \citenamefont {Sheberla}, \citenamefont
  {Aguilera-Iparraguirre}, \citenamefont {Hirzel}, \citenamefont {Adams},\ and\
  \citenamefont {Aspuru-Guzik}}]{gomez2018automatic}%
  \BibitemOpen
  \bibfield  {author} {\bibinfo {author} {\bibfnamefont {R.}~\bibnamefont
  {G{\'o}mez-Bombarelli}}, \bibinfo {author} {\bibfnamefont {J.~N.}\
  \bibnamefont {Wei}}, \bibinfo {author} {\bibfnamefont {D.}~\bibnamefont
  {Duvenaud}}, \bibinfo {author} {\bibfnamefont {J.~M.}\ \bibnamefont
  {Hern{\'a}ndez-Lobato}}, \bibinfo {author} {\bibfnamefont {B.}~\bibnamefont
  {S{\'a}nchez-Lengeling}}, \bibinfo {author} {\bibfnamefont {D.}~\bibnamefont
  {Sheberla}}, \bibinfo {author} {\bibfnamefont {J.}~\bibnamefont
  {Aguilera-Iparraguirre}}, \bibinfo {author} {\bibfnamefont {T.~D.}\
  \bibnamefont {Hirzel}}, \bibinfo {author} {\bibfnamefont {R.~P.}\
  \bibnamefont {Adams}}, \ and\ \bibinfo {author} {\bibfnamefont
  {A.}~\bibnamefont {Aspuru-Guzik}},\ }\href@noop {} {\bibfield  {journal}
  {\bibinfo  {journal} {ACS Cent. Sci.}\ }\textbf {\bibinfo {volume} {4}},\
  \bibinfo {pages} {268} (\bibinfo {year} {2018})}\BibitemShut {NoStop}%
\bibitem [{\citenamefont {Carbone}\ \emph {et~al.}(2019)\citenamefont
  {Carbone}, \citenamefont {Yoo}, \citenamefont {Topsakal},\ and\ \citenamefont
  {Lu}}]{carbone2019classification}%
  \BibitemOpen
  \bibfield  {author} {\bibinfo {author} {\bibfnamefont {M.~R.}\ \bibnamefont
  {Carbone}}, \bibinfo {author} {\bibfnamefont {S.}~\bibnamefont {Yoo}},
  \bibinfo {author} {\bibfnamefont {M.}~\bibnamefont {Topsakal}}, \ and\
  \bibinfo {author} {\bibfnamefont {D.}~\bibnamefont {Lu}},\ }\href@noop {}
  {\bibfield  {journal} {\bibinfo  {journal} {Phys. Rev. Mater.}\ }\textbf
  {\bibinfo {volume} {3}},\ \bibinfo {pages} {033604} (\bibinfo {year}
  {2019})}\BibitemShut {NoStop}%
\bibitem [{\citenamefont {Liu}\ \emph {et~al.}(2019)\citenamefont {Liu},
  \citenamefont {Tao}, \citenamefont {Hsu}, \citenamefont {Du},\ and\
  \citenamefont {Billinge}}]{liu2019using}%
  \BibitemOpen
  \bibfield  {author} {\bibinfo {author} {\bibfnamefont {C.-H.}\ \bibnamefont
  {Liu}}, \bibinfo {author} {\bibfnamefont {Y.}~\bibnamefont {Tao}}, \bibinfo
  {author} {\bibfnamefont {D.}~\bibnamefont {Hsu}}, \bibinfo {author}
  {\bibfnamefont {Q.}~\bibnamefont {Du}}, \ and\ \bibinfo {author}
  {\bibfnamefont {S.~J.}\ \bibnamefont {Billinge}},\ }\href@noop {} {\bibfield
  {journal} {\bibinfo  {journal} {Acta Crystallogr. A}\ }\textbf {\bibinfo
  {volume} {75}},\ \bibinfo {pages} {633} (\bibinfo {year} {2019})}\BibitemShut
  {NoStop}%
\bibitem [{\citenamefont {Zhang}\ \emph {et~al.}(2019)\citenamefont {Zhang},
  \citenamefont {He}, \citenamefont {Chen}, \citenamefont {Bai}, \citenamefont
  {Nolan}, \citenamefont {Roberts}, \citenamefont {Banerjee}, \citenamefont
  {Matsunaga}, \citenamefont {Mo},\ and\ \citenamefont
  {Ling}}]{zhang2019unsupervised}%
  \BibitemOpen
  \bibfield  {author} {\bibinfo {author} {\bibfnamefont {Y.}~\bibnamefont
  {Zhang}}, \bibinfo {author} {\bibfnamefont {X.}~\bibnamefont {He}}, \bibinfo
  {author} {\bibfnamefont {Z.}~\bibnamefont {Chen}}, \bibinfo {author}
  {\bibfnamefont {Q.}~\bibnamefont {Bai}}, \bibinfo {author} {\bibfnamefont
  {A.~M.}\ \bibnamefont {Nolan}}, \bibinfo {author} {\bibfnamefont {C.~A.}\
  \bibnamefont {Roberts}}, \bibinfo {author} {\bibfnamefont {D.}~\bibnamefont
  {Banerjee}}, \bibinfo {author} {\bibfnamefont {T.}~\bibnamefont {Matsunaga}},
  \bibinfo {author} {\bibfnamefont {Y.}~\bibnamefont {Mo}}, \ and\ \bibinfo
  {author} {\bibfnamefont {C.}~\bibnamefont {Ling}},\ }\href@noop {} {\bibfield
   {journal} {\bibinfo  {journal} {Nat. Comm.}\ }\textbf {\bibinfo {volume}
  {10}},\ \bibinfo {pages} {1} (\bibinfo {year} {2019})}\BibitemShut {NoStop}%
\bibitem [{\citenamefont {Torrisi}\ \emph {et~al.}(2020)\citenamefont
  {Torrisi}, \citenamefont {Carbone}, \citenamefont {Rohr}, \citenamefont
  {Montoya}, \citenamefont {Ha}, \citenamefont {Yano}, \citenamefont {Suram},\
  and\ \citenamefont {Hung}}]{torrisi2020random}%
  \BibitemOpen
  \bibfield  {author} {\bibinfo {author} {\bibfnamefont {S.~B.}\ \bibnamefont
  {Torrisi}}, \bibinfo {author} {\bibfnamefont {M.~R.}\ \bibnamefont
  {Carbone}}, \bibinfo {author} {\bibfnamefont {B.~A.}\ \bibnamefont {Rohr}},
  \bibinfo {author} {\bibfnamefont {J.~H.}\ \bibnamefont {Montoya}}, \bibinfo
  {author} {\bibfnamefont {Y.}~\bibnamefont {Ha}}, \bibinfo {author}
  {\bibfnamefont {J.}~\bibnamefont {Yano}}, \bibinfo {author} {\bibfnamefont
  {S.~K.}\ \bibnamefont {Suram}}, \ and\ \bibinfo {author} {\bibfnamefont
  {L.}~\bibnamefont {Hung}},\ }\href@noop {} {\bibfield  {journal} {\bibinfo
  {journal} {npj Comput. Mater.}\ }\textbf {\bibinfo {volume} {6}},\ \bibinfo
  {pages} {1} (\bibinfo {year} {2020})}\BibitemShut {NoStop}%
\bibitem [{\citenamefont {Mercado}\ \emph {et~al.}(2021)\citenamefont
  {Mercado}, \citenamefont {Rastemo}, \citenamefont {Lindel{\"o}f},
  \citenamefont {Klambauer}, \citenamefont {Engkvist}, \citenamefont {Chen},\
  and\ \citenamefont {Bjerrum}}]{mercado2021graph}%
  \BibitemOpen
  \bibfield  {author} {\bibinfo {author} {\bibfnamefont {R.}~\bibnamefont
  {Mercado}}, \bibinfo {author} {\bibfnamefont {T.}~\bibnamefont {Rastemo}},
  \bibinfo {author} {\bibfnamefont {E.}~\bibnamefont {Lindel{\"o}f}}, \bibinfo
  {author} {\bibfnamefont {G.}~\bibnamefont {Klambauer}}, \bibinfo {author}
  {\bibfnamefont {O.}~\bibnamefont {Engkvist}}, \bibinfo {author}
  {\bibfnamefont {H.}~\bibnamefont {Chen}}, \ and\ \bibinfo {author}
  {\bibfnamefont {E.~J.}\ \bibnamefont {Bjerrum}},\ }\href@noop {} {\bibfield
  {journal} {\bibinfo  {journal} {Mach. Learn. Sci. Technol.}\ }\textbf
  {\bibinfo {volume} {2}},\ \bibinfo {pages} {025023} (\bibinfo {year}
  {2021})}\BibitemShut {NoStop}%
\bibitem [{\citenamefont {Mouchlis}\ \emph {et~al.}(2021)\citenamefont
  {Mouchlis}, \citenamefont {Afantitis}, \citenamefont {Serra}, \citenamefont
  {Fratello}, \citenamefont {Papadiamantis}, \citenamefont {Aidinis},
  \citenamefont {Lynch}, \citenamefont {Greco},\ and\ \citenamefont
  {Melagraki}}]{mouchlis2021advances}%
  \BibitemOpen
  \bibfield  {author} {\bibinfo {author} {\bibfnamefont {V.~D.}\ \bibnamefont
  {Mouchlis}}, \bibinfo {author} {\bibfnamefont {A.}~\bibnamefont {Afantitis}},
  \bibinfo {author} {\bibfnamefont {A.}~\bibnamefont {Serra}}, \bibinfo
  {author} {\bibfnamefont {M.}~\bibnamefont {Fratello}}, \bibinfo {author}
  {\bibfnamefont {A.~G.}\ \bibnamefont {Papadiamantis}}, \bibinfo {author}
  {\bibfnamefont {V.}~\bibnamefont {Aidinis}}, \bibinfo {author} {\bibfnamefont
  {I.}~\bibnamefont {Lynch}}, \bibinfo {author} {\bibfnamefont
  {D.}~\bibnamefont {Greco}}, \ and\ \bibinfo {author} {\bibfnamefont
  {G.}~\bibnamefont {Melagraki}},\ }\href@noop {} {\bibfield  {journal}
  {\bibinfo  {journal} {Int. J. Mol. Sci.}\ }\textbf {\bibinfo {volume} {22}},\
  \bibinfo {pages} {1676} (\bibinfo {year} {2021})}\BibitemShut {NoStop}%
\bibitem [{\citenamefont {Gilmer}\ \emph {et~al.}(2017)\citenamefont {Gilmer},
  \citenamefont {Schoenholz}, \citenamefont {Riley}, \citenamefont {Vinyals},\
  and\ \citenamefont {Dahl}}]{gilmer2017neural}%
  \BibitemOpen
  \bibfield  {author} {\bibinfo {author} {\bibfnamefont {J.}~\bibnamefont
  {Gilmer}}, \bibinfo {author} {\bibfnamefont {S.~S.}\ \bibnamefont
  {Schoenholz}}, \bibinfo {author} {\bibfnamefont {P.~F.}\ \bibnamefont
  {Riley}}, \bibinfo {author} {\bibfnamefont {O.}~\bibnamefont {Vinyals}}, \
  and\ \bibinfo {author} {\bibfnamefont {G.~E.}\ \bibnamefont {Dahl}},\ }in\
  \href@noop {} {\emph {\bibinfo {booktitle} {International conference on
  machine learning}}}\ (\bibinfo {organization} {PMLR},\ \bibinfo {year}
  {2017})\ pp.\ \bibinfo {pages} {1263--1272}\BibitemShut {NoStop}%
\bibitem [{\citenamefont {Xie}\ and\ \citenamefont
  {Grossman}(2018)}]{xie2018crystal}%
  \BibitemOpen
  \bibfield  {author} {\bibinfo {author} {\bibfnamefont {T.}~\bibnamefont
  {Xie}}\ and\ \bibinfo {author} {\bibfnamefont {J.~C.}\ \bibnamefont
  {Grossman}},\ }\href@noop {} {\bibfield  {journal} {\bibinfo  {journal}
  {Phys. Rev. Lett.}\ }\textbf {\bibinfo {volume} {120}},\ \bibinfo {pages}
  {145301} (\bibinfo {year} {2018})}\BibitemShut {NoStop}%
\bibitem [{\citenamefont {Carbone}\ \emph {et~al.}(2020)\citenamefont
  {Carbone}, \citenamefont {Topsakal}, \citenamefont {Lu},\ and\ \citenamefont
  {Yoo}}]{carbone2020machine}%
  \BibitemOpen
  \bibfield  {author} {\bibinfo {author} {\bibfnamefont {M.~R.}\ \bibnamefont
  {Carbone}}, \bibinfo {author} {\bibfnamefont {M.}~\bibnamefont {Topsakal}},
  \bibinfo {author} {\bibfnamefont {D.}~\bibnamefont {Lu}}, \ and\ \bibinfo
  {author} {\bibfnamefont {S.}~\bibnamefont {Yoo}},\ }\href@noop {} {\bibfield
  {journal} {\bibinfo  {journal} {Phys. Rev. Lett.}\ }\textbf {\bibinfo
  {volume} {124}},\ \bibinfo {pages} {156401} (\bibinfo {year}
  {2020})}\BibitemShut {NoStop}%
\bibitem [{\citenamefont {Rankine}\ and\ \citenamefont
  {Penfold}(2022)}]{rankine2022accurate}%
  \BibitemOpen
  \bibfield  {author} {\bibinfo {author} {\bibfnamefont {C.~D.}\ \bibnamefont
  {Rankine}}\ and\ \bibinfo {author} {\bibfnamefont {T.}~\bibnamefont
  {Penfold}},\ }\href@noop {} {\bibfield  {journal} {\bibinfo  {journal} {J.
  Chem. Phys.}\ }\textbf {\bibinfo {volume} {156}},\ \bibinfo {pages} {164102}
  (\bibinfo {year} {2022})}\BibitemShut {NoStop}%
\bibitem [{\citenamefont {Noack}\ \emph {et~al.}(2020)\citenamefont {Noack},
  \citenamefont {Doerk}, \citenamefont {Li}, \citenamefont {Fukuto},\ and\
  \citenamefont {Yager}}]{noack2020advances}%
  \BibitemOpen
  \bibfield  {author} {\bibinfo {author} {\bibfnamefont {M.~M.}\ \bibnamefont
  {Noack}}, \bibinfo {author} {\bibfnamefont {G.~S.}\ \bibnamefont {Doerk}},
  \bibinfo {author} {\bibfnamefont {R.}~\bibnamefont {Li}}, \bibinfo {author}
  {\bibfnamefont {M.}~\bibnamefont {Fukuto}}, \ and\ \bibinfo {author}
  {\bibfnamefont {K.~G.}\ \bibnamefont {Yager}},\ }\href@noop {} {\bibfield
  {journal} {\bibinfo  {journal} {Sci. Rep.}\ }\textbf {\bibinfo {volume}
  {10}},\ \bibinfo {pages} {1} (\bibinfo {year} {2020})}\BibitemShut {NoStop}%
\bibitem [{\citenamefont {MacLeod}\ \emph {et~al.}(2020)\citenamefont
  {MacLeod}, \citenamefont {Parlane}, \citenamefont {Morrissey}, \citenamefont
  {H{\"a}se}, \citenamefont {Roch}, \citenamefont {Dettelbach}, \citenamefont
  {Moreira}, \citenamefont {Yunker}, \citenamefont {Rooney}, \citenamefont
  {Deeth} \emph {et~al.}}]{macleod2020self}%
  \BibitemOpen
  \bibfield  {author} {\bibinfo {author} {\bibfnamefont {B.~P.}\ \bibnamefont
  {MacLeod}}, \bibinfo {author} {\bibfnamefont {F.~G.}\ \bibnamefont
  {Parlane}}, \bibinfo {author} {\bibfnamefont {T.~D.}\ \bibnamefont
  {Morrissey}}, \bibinfo {author} {\bibfnamefont {F.}~\bibnamefont {H{\"a}se}},
  \bibinfo {author} {\bibfnamefont {L.~M.}\ \bibnamefont {Roch}}, \bibinfo
  {author} {\bibfnamefont {K.~E.}\ \bibnamefont {Dettelbach}}, \bibinfo
  {author} {\bibfnamefont {R.}~\bibnamefont {Moreira}}, \bibinfo {author}
  {\bibfnamefont {L.~P.}\ \bibnamefont {Yunker}}, \bibinfo {author}
  {\bibfnamefont {M.~B.}\ \bibnamefont {Rooney}}, \bibinfo {author}
  {\bibfnamefont {J.~R.}\ \bibnamefont {Deeth}},  \emph {et~al.},\ }\href@noop
  {} {\bibfield  {journal} {\bibinfo  {journal} {Sci. Adv.}\ }\textbf {\bibinfo
  {volume} {6}},\ \bibinfo {pages} {eaaz8867} (\bibinfo {year}
  {2020})}\BibitemShut {NoStop}%
\bibitem [{\citenamefont {Epps}\ \emph {et~al.}(2020)\citenamefont {Epps},
  \citenamefont {Bowen}, \citenamefont {Volk}, \citenamefont {Abdel-Latif},
  \citenamefont {Han}, \citenamefont {Reyes}, \citenamefont {Amassian},\ and\
  \citenamefont {Abolhasani}}]{epps2020artificial}%
  \BibitemOpen
  \bibfield  {author} {\bibinfo {author} {\bibfnamefont {R.~W.}\ \bibnamefont
  {Epps}}, \bibinfo {author} {\bibfnamefont {M.~S.}\ \bibnamefont {Bowen}},
  \bibinfo {author} {\bibfnamefont {A.~A.}\ \bibnamefont {Volk}}, \bibinfo
  {author} {\bibfnamefont {K.}~\bibnamefont {Abdel-Latif}}, \bibinfo {author}
  {\bibfnamefont {S.}~\bibnamefont {Han}}, \bibinfo {author} {\bibfnamefont
  {K.~G.}\ \bibnamefont {Reyes}}, \bibinfo {author} {\bibfnamefont
  {A.}~\bibnamefont {Amassian}}, \ and\ \bibinfo {author} {\bibfnamefont
  {M.}~\bibnamefont {Abolhasani}},\ }\href@noop {} {\bibfield  {journal}
  {\bibinfo  {journal} {Adv. Mater.}\ }\textbf {\bibinfo {volume} {32}},\
  \bibinfo {pages} {2001626} (\bibinfo {year} {2020})}\BibitemShut {NoStop}%
\bibitem [{\citenamefont {Noack}\ \emph {et~al.}(2021)\citenamefont {Noack},
  \citenamefont {Zwart}, \citenamefont {Ushizima}, \citenamefont {Fukuto},
  \citenamefont {Yager}, \citenamefont {Elbert}, \citenamefont {Murray},
  \citenamefont {Stein}, \citenamefont {Doerk}, \citenamefont {Tsai} \emph
  {et~al.}}]{noack2021gaussian}%
  \BibitemOpen
  \bibfield  {author} {\bibinfo {author} {\bibfnamefont {M.~M.}\ \bibnamefont
  {Noack}}, \bibinfo {author} {\bibfnamefont {P.~H.}\ \bibnamefont {Zwart}},
  \bibinfo {author} {\bibfnamefont {D.~M.}\ \bibnamefont {Ushizima}}, \bibinfo
  {author} {\bibfnamefont {M.}~\bibnamefont {Fukuto}}, \bibinfo {author}
  {\bibfnamefont {K.~G.}\ \bibnamefont {Yager}}, \bibinfo {author}
  {\bibfnamefont {K.~C.}\ \bibnamefont {Elbert}}, \bibinfo {author}
  {\bibfnamefont {C.~B.}\ \bibnamefont {Murray}}, \bibinfo {author}
  {\bibfnamefont {A.}~\bibnamefont {Stein}}, \bibinfo {author} {\bibfnamefont
  {G.~S.}\ \bibnamefont {Doerk}}, \bibinfo {author} {\bibfnamefont {E.~H.}\
  \bibnamefont {Tsai}},  \emph {et~al.},\ }\href@noop {} {\bibfield  {journal}
  {\bibinfo  {journal} {Nat. Rev. Phys.}\ }\textbf {\bibinfo {volume} {3}},\
  \bibinfo {pages} {685} (\bibinfo {year} {2021})}\BibitemShut {NoStop}%
\bibitem [{\citenamefont {Bateni}\ \emph {et~al.}(2022)\citenamefont {Bateni},
  \citenamefont {Epps}, \citenamefont {Antami}, \citenamefont {Dargis},
  \citenamefont {Bennett}, \citenamefont {Reyes},\ and\ \citenamefont
  {Abolhasani}}]{bateni2022autonomous}%
  \BibitemOpen
  \bibfield  {author} {\bibinfo {author} {\bibfnamefont {F.}~\bibnamefont
  {Bateni}}, \bibinfo {author} {\bibfnamefont {R.~W.}\ \bibnamefont {Epps}},
  \bibinfo {author} {\bibfnamefont {K.}~\bibnamefont {Antami}}, \bibinfo
  {author} {\bibfnamefont {R.}~\bibnamefont {Dargis}}, \bibinfo {author}
  {\bibfnamefont {J.~A.}\ \bibnamefont {Bennett}}, \bibinfo {author}
  {\bibfnamefont {K.~G.}\ \bibnamefont {Reyes}}, \ and\ \bibinfo {author}
  {\bibfnamefont {M.}~\bibnamefont {Abolhasani}},\ }\href@noop {} {\bibfield
  {journal} {\bibinfo  {journal} {Adv. Intell. Syst.}\ ,\ \bibinfo {pages}
  {2200017}} (\bibinfo {year} {2022})}\BibitemShut {NoStop}%
\bibitem [{\citenamefont {Konstantinova}\ \emph {et~al.}(2022)\citenamefont
  {Konstantinova}, \citenamefont {Maffettone}, \citenamefont {Ravel},
  \citenamefont {Campbell}, \citenamefont {Barbour},\ and\ \citenamefont
  {Olds}}]{D2DD00014H}%
  \BibitemOpen
  \bibfield  {author} {\bibinfo {author} {\bibfnamefont {T.}~\bibnamefont
  {Konstantinova}}, \bibinfo {author} {\bibfnamefont {P.~M.}\ \bibnamefont
  {Maffettone}}, \bibinfo {author} {\bibfnamefont {B.}~\bibnamefont {Ravel}},
  \bibinfo {author} {\bibfnamefont {S.~I.}\ \bibnamefont {Campbell}}, \bibinfo
  {author} {\bibfnamefont {A.~M.}\ \bibnamefont {Barbour}}, \ and\ \bibinfo
  {author} {\bibfnamefont {D.}~\bibnamefont {Olds}},\ }\href@noop {} {\bibfield
   {journal} {\bibinfo  {journal} {Digital Discovery}\ ,\ } (\bibinfo {year}
  {2022})}\BibitemShut {NoStop}%
\bibitem [{\citenamefont {Behler}\ and\ \citenamefont
  {Parrinello}(2007)}]{behler2007generalized}%
  \BibitemOpen
  \bibfield  {author} {\bibinfo {author} {\bibfnamefont {J.}~\bibnamefont
  {Behler}}\ and\ \bibinfo {author} {\bibfnamefont {M.}~\bibnamefont
  {Parrinello}},\ }\href@noop {} {\bibfield  {journal} {\bibinfo  {journal}
  {Phys. Rev. Lett.}\ }\textbf {\bibinfo {volume} {98}},\ \bibinfo {pages}
  {146401} (\bibinfo {year} {2007})}\BibitemShut {NoStop}%
\bibitem [{\citenamefont {Behler}(2011{\natexlab{a}})}]{behler2011atom}%
  \BibitemOpen
  \bibfield  {author} {\bibinfo {author} {\bibfnamefont {J.}~\bibnamefont
  {Behler}},\ }\href@noop {} {\bibfield  {journal} {\bibinfo  {journal} {J.
  Chem. Phys.}\ }\textbf {\bibinfo {volume} {134}},\ \bibinfo {pages} {074106}
  (\bibinfo {year} {2011}{\natexlab{a}})}\BibitemShut {NoStop}%
\bibitem [{\citenamefont {Behler}(2011{\natexlab{b}})}]{behler2011neural}%
  \BibitemOpen
  \bibfield  {author} {\bibinfo {author} {\bibfnamefont {J.}~\bibnamefont
  {Behler}},\ }\href@noop {} {\bibfield  {journal} {\bibinfo  {journal} {Phys.
  Chem. Chem. Phys.}\ }\textbf {\bibinfo {volume} {13}},\ \bibinfo {pages}
  {17930} (\bibinfo {year} {2011}{\natexlab{b}})}\BibitemShut {NoStop}%
\bibitem [{\citenamefont {Artrith}\ and\ \citenamefont
  {Urban}(2016)}]{artrith2016implementation}%
  \BibitemOpen
  \bibfield  {author} {\bibinfo {author} {\bibfnamefont {N.}~\bibnamefont
  {Artrith}}\ and\ \bibinfo {author} {\bibfnamefont {A.}~\bibnamefont
  {Urban}},\ }\href@noop {} {\bibfield  {journal} {\bibinfo  {journal} {Comput.
  Mater. Sci.}\ }\textbf {\bibinfo {volume} {114}},\ \bibinfo {pages} {135}
  (\bibinfo {year} {2016})}\BibitemShut {NoStop}%
\bibitem [{\citenamefont {Behler}(2021)}]{behler2021four}%
  \BibitemOpen
  \bibfield  {author} {\bibinfo {author} {\bibfnamefont {J.}~\bibnamefont
  {Behler}},\ }\href@noop {} {\bibfield  {journal} {\bibinfo  {journal} {Chem.
  Rev.}\ }\textbf {\bibinfo {volume} {121}},\ \bibinfo {pages} {10037}
  (\bibinfo {year} {2021})}\BibitemShut {NoStop}%
\bibitem [{\citenamefont {Gomes}\ \emph {et~al.}(2019)\citenamefont {Gomes},
  \citenamefont {Selman},\ and\ \citenamefont
  {Gregoire}}]{gomes2019artificial}%
  \BibitemOpen
  \bibfield  {author} {\bibinfo {author} {\bibfnamefont {C.~P.}\ \bibnamefont
  {Gomes}}, \bibinfo {author} {\bibfnamefont {B.}~\bibnamefont {Selman}}, \
  and\ \bibinfo {author} {\bibfnamefont {J.~M.}\ \bibnamefont {Gregoire}},\
  }\href@noop {} {\bibfield  {journal} {\bibinfo  {journal} {MRS Bull.}\
  }\textbf {\bibinfo {volume} {44}},\ \bibinfo {pages} {538} (\bibinfo {year}
  {2019})}\BibitemShut {NoStop}%
\bibitem [{\citenamefont {Wei}\ \emph {et~al.}(2019)\citenamefont {Wei},
  \citenamefont {Chu}, \citenamefont {Sun}, \citenamefont {Xu}, \citenamefont
  {Deng}, \citenamefont {Chen}, \citenamefont {Wei},\ and\ \citenamefont
  {Lei}}]{wei2019machine}%
  \BibitemOpen
  \bibfield  {author} {\bibinfo {author} {\bibfnamefont {J.}~\bibnamefont
  {Wei}}, \bibinfo {author} {\bibfnamefont {X.}~\bibnamefont {Chu}}, \bibinfo
  {author} {\bibfnamefont {X.-Y.}\ \bibnamefont {Sun}}, \bibinfo {author}
  {\bibfnamefont {K.}~\bibnamefont {Xu}}, \bibinfo {author} {\bibfnamefont
  {H.-X.}\ \bibnamefont {Deng}}, \bibinfo {author} {\bibfnamefont
  {J.}~\bibnamefont {Chen}}, \bibinfo {author} {\bibfnamefont {Z.}~\bibnamefont
  {Wei}}, \ and\ \bibinfo {author} {\bibfnamefont {M.}~\bibnamefont {Lei}},\
  }\href@noop {} {\bibfield  {journal} {\bibinfo  {journal} {InfoMat}\ }\textbf
  {\bibinfo {volume} {1}},\ \bibinfo {pages} {338} (\bibinfo {year}
  {2019})}\BibitemShut {NoStop}%
\bibitem [{\citenamefont {Morgan}\ and\ \citenamefont
  {Jacobs}(2020)}]{morgan2020opportunities}%
  \BibitemOpen
  \bibfield  {author} {\bibinfo {author} {\bibfnamefont {D.}~\bibnamefont
  {Morgan}}\ and\ \bibinfo {author} {\bibfnamefont {R.}~\bibnamefont
  {Jacobs}},\ }\href@noop {} {\bibfield  {journal} {\bibinfo  {journal} {Annu.
  Rev. Mater. Res.}\ }\textbf {\bibinfo {volume} {50}},\ \bibinfo {pages} {71}
  (\bibinfo {year} {2020})}\BibitemShut {NoStop}%
\bibitem [{\citenamefont {Wang}\ \emph {et~al.}(2020)\citenamefont {Wang},
  \citenamefont {Murdock}, \citenamefont {Kauwe}, \citenamefont {Oliynyk},
  \citenamefont {Gurlo}, \citenamefont {Brgoch}, \citenamefont {Persson},\ and\
  \citenamefont {Sparks}}]{wang2020machine}%
  \BibitemOpen
  \bibfield  {author} {\bibinfo {author} {\bibfnamefont {A.~Y.-T.}\
  \bibnamefont {Wang}}, \bibinfo {author} {\bibfnamefont {R.~J.}\ \bibnamefont
  {Murdock}}, \bibinfo {author} {\bibfnamefont {S.~K.}\ \bibnamefont {Kauwe}},
  \bibinfo {author} {\bibfnamefont {A.~O.}\ \bibnamefont {Oliynyk}}, \bibinfo
  {author} {\bibfnamefont {A.}~\bibnamefont {Gurlo}}, \bibinfo {author}
  {\bibfnamefont {J.}~\bibnamefont {Brgoch}}, \bibinfo {author} {\bibfnamefont
  {K.~A.}\ \bibnamefont {Persson}}, \ and\ \bibinfo {author} {\bibfnamefont
  {T.~D.}\ \bibnamefont {Sparks}},\ }\href@noop {} {\bibfield  {journal}
  {\bibinfo  {journal} {Chem. Mater.}\ }\textbf {\bibinfo {volume} {32}},\
  \bibinfo {pages} {4954} (\bibinfo {year} {2020})}\BibitemShut {NoStop}%
\bibitem [{\citenamefont {Artrith}\ \emph {et~al.}(2021)\citenamefont
  {Artrith}, \citenamefont {Butler}, \citenamefont {Coudert}, \citenamefont
  {Han}, \citenamefont {Isayev}, \citenamefont {Jain},\ and\ \citenamefont
  {Walsh}}]{artrith2021best}%
  \BibitemOpen
  \bibfield  {author} {\bibinfo {author} {\bibfnamefont {N.}~\bibnamefont
  {Artrith}}, \bibinfo {author} {\bibfnamefont {K.~T.}\ \bibnamefont {Butler}},
  \bibinfo {author} {\bibfnamefont {F.-X.}\ \bibnamefont {Coudert}}, \bibinfo
  {author} {\bibfnamefont {S.}~\bibnamefont {Han}}, \bibinfo {author}
  {\bibfnamefont {O.}~\bibnamefont {Isayev}}, \bibinfo {author} {\bibfnamefont
  {A.}~\bibnamefont {Jain}}, \ and\ \bibinfo {author} {\bibfnamefont
  {A.}~\bibnamefont {Walsh}},\ }\href@noop {} {\bibfield  {journal} {\bibinfo
  {journal} {Nat. Chem.}\ }\textbf {\bibinfo {volume} {13}},\ \bibinfo {pages}
  {505} (\bibinfo {year} {2021})}\BibitemShut {NoStop}%
\bibitem [{\citenamefont {Sambasivan}\ \emph {et~al.}(2021)\citenamefont
  {Sambasivan}, \citenamefont {Kapania}, \citenamefont {Highfill},
  \citenamefont {Akrong}, \citenamefont {Paritosh},\ and\ \citenamefont
  {Aroyo}}]{sambasivan2021everyone}%
  \BibitemOpen
  \bibfield  {author} {\bibinfo {author} {\bibfnamefont {N.}~\bibnamefont
  {Sambasivan}}, \bibinfo {author} {\bibfnamefont {S.}~\bibnamefont {Kapania}},
  \bibinfo {author} {\bibfnamefont {H.}~\bibnamefont {Highfill}}, \bibinfo
  {author} {\bibfnamefont {D.}~\bibnamefont {Akrong}}, \bibinfo {author}
  {\bibfnamefont {P.}~\bibnamefont {Paritosh}}, \ and\ \bibinfo {author}
  {\bibfnamefont {L.~M.}\ \bibnamefont {Aroyo}},\ }in\ \href@noop {} {\emph
  {\bibinfo {booktitle} {proceedings of the 2021 CHI Conference on Human
  Factors in Computing Systems}}}\ (\bibinfo {year} {2021})\ pp.\ \bibinfo
  {pages} {1--15}\BibitemShut {NoStop}%
\bibitem [{\citenamefont {Chollet}(2021)}]{chollet2021deep}%
  \BibitemOpen
  \bibfield  {author} {\bibinfo {author} {\bibfnamefont {F.}~\bibnamefont
  {Chollet}},\ }\href@noop {} {\emph {\bibinfo {title} {Deep learning with
  Python}}}\ (\bibinfo  {publisher} {Simon and Schuster},\ \bibinfo {year}
  {2021})\BibitemShut {NoStop}%
\bibitem [{\citenamefont {Ayodele}(2010)}]{ayodele2010types}%
  \BibitemOpen
  \bibfield  {author} {\bibinfo {author} {\bibfnamefont {T.~O.}\ \bibnamefont
  {Ayodele}},\ }\href@noop {} {\bibfield  {journal} {\bibinfo  {journal} {New
  Advances in Machine Learning}\ }\textbf {\bibinfo {volume} {3}},\ \bibinfo
  {pages} {19} (\bibinfo {year} {2010})}\BibitemShut {NoStop}%
\bibitem [{\citenamefont {Mnih}\ \emph {et~al.}(2015)\citenamefont {Mnih},
  \citenamefont {Kavukcuoglu}, \citenamefont {Silver}, \citenamefont {Rusu},
  \citenamefont {Veness}, \citenamefont {Bellemare}, \citenamefont {Graves},
  \citenamefont {Riedmiller}, \citenamefont {Fidjeland}, \citenamefont
  {Ostrovski} \emph {et~al.}}]{mnih2015human}%
  \BibitemOpen
  \bibfield  {author} {\bibinfo {author} {\bibfnamefont {V.}~\bibnamefont
  {Mnih}}, \bibinfo {author} {\bibfnamefont {K.}~\bibnamefont {Kavukcuoglu}},
  \bibinfo {author} {\bibfnamefont {D.}~\bibnamefont {Silver}}, \bibinfo
  {author} {\bibfnamefont {A.~A.}\ \bibnamefont {Rusu}}, \bibinfo {author}
  {\bibfnamefont {J.}~\bibnamefont {Veness}}, \bibinfo {author} {\bibfnamefont
  {M.~G.}\ \bibnamefont {Bellemare}}, \bibinfo {author} {\bibfnamefont
  {A.}~\bibnamefont {Graves}}, \bibinfo {author} {\bibfnamefont
  {M.}~\bibnamefont {Riedmiller}}, \bibinfo {author} {\bibfnamefont {A.~K.}\
  \bibnamefont {Fidjeland}}, \bibinfo {author} {\bibfnamefont {G.}~\bibnamefont
  {Ostrovski}},  \emph {et~al.},\ }\href@noop {} {\bibfield  {journal}
  {\bibinfo  {journal} {Nature}\ }\textbf {\bibinfo {volume} {518}},\ \bibinfo
  {pages} {529} (\bibinfo {year} {2015})}\BibitemShut {NoStop}%
\bibitem [{\citenamefont {Kingma}\ and\ \citenamefont
  {Welling}(2014)}]{kingma_auto-encoding_2014}%
  \BibitemOpen
  \bibfield  {author} {\bibinfo {author} {\bibfnamefont {D.~P.}\ \bibnamefont
  {Kingma}}\ and\ \bibinfo {author} {\bibfnamefont {M.}~\bibnamefont
  {Welling}},\ }in\ \href@noop {} {\emph {\bibinfo {booktitle} {2nd
  International Conference on Learning Representations, {ICLR} 2014, Banff, AB,
  Canada, April 14-16, 2014, Conference Track Proceedings}}},\ \bibinfo
  {editor} {edited by\ \bibinfo {editor} {\bibfnamefont {Y.}~\bibnamefont
  {Bengio}}\ and\ \bibinfo {editor} {\bibfnamefont {Y.}~\bibnamefont {LeCun}}}\
  (\bibinfo {year} {2014})\BibitemShut {NoStop}%
\bibitem [{\citenamefont {Higgins}\ \emph {et~al.}(2017)\citenamefont
  {Higgins}, \citenamefont {Matthey}, \citenamefont {Pal}, \citenamefont
  {Burgess}, \citenamefont {Glorot}, \citenamefont {Botvinick}, \citenamefont
  {Mohamed},\ and\ \citenamefont {Lerchner}}]{higgins2016beta}%
  \BibitemOpen
  \bibfield  {author} {\bibinfo {author} {\bibfnamefont {I.}~\bibnamefont
  {Higgins}}, \bibinfo {author} {\bibfnamefont {L.}~\bibnamefont {Matthey}},
  \bibinfo {author} {\bibfnamefont {A.}~\bibnamefont {Pal}}, \bibinfo {author}
  {\bibfnamefont {C.}~\bibnamefont {Burgess}}, \bibinfo {author} {\bibfnamefont
  {X.}~\bibnamefont {Glorot}}, \bibinfo {author} {\bibfnamefont
  {M.}~\bibnamefont {Botvinick}}, \bibinfo {author} {\bibfnamefont
  {S.}~\bibnamefont {Mohamed}}, \ and\ \bibinfo {author} {\bibfnamefont
  {A.}~\bibnamefont {Lerchner}},\ }in\ \href@noop {} {\emph {\bibinfo
  {booktitle} {5th International Conference on Learning Representations, {ICLR}
  2017, Toulon, France, April 24-26, 2017, Conference Track Proceedings}}}\
  (\bibinfo  {publisher} {OpenReview.net},\ \bibinfo {year} {2017})\BibitemShut
  {NoStop}%
\bibitem [{\citenamefont {Miles}\ \emph {et~al.}(2021)\citenamefont {Miles},
  \citenamefont {Carbone}, \citenamefont {Sturm}, \citenamefont {Lu},
  \citenamefont {Weichselbaum}, \citenamefont {Barros},\ and\ \citenamefont
  {Konik}}]{miles2021machine}%
  \BibitemOpen
  \bibfield  {author} {\bibinfo {author} {\bibfnamefont {C.}~\bibnamefont
  {Miles}}, \bibinfo {author} {\bibfnamefont {M.~R.}\ \bibnamefont {Carbone}},
  \bibinfo {author} {\bibfnamefont {E.~J.}\ \bibnamefont {Sturm}}, \bibinfo
  {author} {\bibfnamefont {D.}~\bibnamefont {Lu}}, \bibinfo {author}
  {\bibfnamefont {A.}~\bibnamefont {Weichselbaum}}, \bibinfo {author}
  {\bibfnamefont {K.}~\bibnamefont {Barros}}, \ and\ \bibinfo {author}
  {\bibfnamefont {R.~M.}\ \bibnamefont {Konik}},\ }\href@noop {} {\bibfield
  {journal} {\bibinfo  {journal} {Phys. Rev. B.}\ }\textbf {\bibinfo {volume}
  {104}},\ \bibinfo {pages} {235111} (\bibinfo {year} {2021})}\BibitemShut
  {NoStop}%
\bibitem [{\citenamefont {Kingma}\ and\ \citenamefont
  {Ba}(2014)}]{kingma2014adam}%
  \BibitemOpen
  \bibfield  {author} {\bibinfo {author} {\bibfnamefont {D.~P.}\ \bibnamefont
  {Kingma}}\ and\ \bibinfo {author} {\bibfnamefont {J.}~\bibnamefont {Ba}},\
  }\href@noop {} {\bibfield  {journal} {\bibinfo  {journal} {arXiv preprint
  arXiv:1412.6980}\ } (\bibinfo {year} {2014})}\BibitemShut {NoStop}%
\bibitem [{\citenamefont {Weininger}(1988)}]{weininger1988smiles}%
  \BibitemOpen
  \bibfield  {author} {\bibinfo {author} {\bibfnamefont {D.}~\bibnamefont
  {Weininger}},\ }\href@noop {} {\bibfield  {journal} {\bibinfo  {journal} {J.
  Chem. Inf. Model.}\ }\textbf {\bibinfo {volume} {28}},\ \bibinfo {pages} {31}
  (\bibinfo {year} {1988})}\BibitemShut {NoStop}%
\bibitem [{\citenamefont {Rasmussen}\ and\ \citenamefont
  {Williams}(2006)}]{Rasmussen2006GP}%
  \BibitemOpen
  \bibfield  {author} {\bibinfo {author} {\bibfnamefont {C.~E.}\ \bibnamefont
  {Rasmussen}}\ and\ \bibinfo {author} {\bibfnamefont {C.~K.~I.}\ \bibnamefont
  {Williams}},\ }\href@noop {} {\emph {\bibinfo {title} {Gaussian Processes for
  Machine Learning}}}\ (\bibinfo  {publisher} {Cambridge: MIT Press},\ \bibinfo
  {year} {2006})\BibitemShut {NoStop}%
\bibitem [{\citenamefont {Jain}\ \emph {et~al.}(2013)\citenamefont {Jain},
  \citenamefont {Ong}, \citenamefont {Hautier}, \citenamefont {Chen},
  \citenamefont {Richards}, \citenamefont {Dacek}, \citenamefont {Cholia},
  \citenamefont {Gunter}, \citenamefont {Skinner}, \citenamefont {Ceder},\ and\
  \citenamefont {Persson}}]{Jain2013}%
  \BibitemOpen
  \bibfield  {author} {\bibinfo {author} {\bibfnamefont {A.}~\bibnamefont
  {Jain}}, \bibinfo {author} {\bibfnamefont {S.~P.}\ \bibnamefont {Ong}},
  \bibinfo {author} {\bibfnamefont {G.}~\bibnamefont {Hautier}}, \bibinfo
  {author} {\bibfnamefont {W.}~\bibnamefont {Chen}}, \bibinfo {author}
  {\bibfnamefont {W.~D.}\ \bibnamefont {Richards}}, \bibinfo {author}
  {\bibfnamefont {S.}~\bibnamefont {Dacek}}, \bibinfo {author} {\bibfnamefont
  {S.}~\bibnamefont {Cholia}}, \bibinfo {author} {\bibfnamefont
  {D.}~\bibnamefont {Gunter}}, \bibinfo {author} {\bibfnamefont
  {D.}~\bibnamefont {Skinner}}, \bibinfo {author} {\bibfnamefont
  {G.}~\bibnamefont {Ceder}}, \ and\ \bibinfo {author} {\bibfnamefont {K.~a.}\
  \bibnamefont {Persson}},\ }\href@noop {} {\bibfield  {journal} {\bibinfo
  {journal} {APL Mater.}\ }\textbf {\bibinfo {volume} {1}},\ \bibinfo {pages}
  {011002} (\bibinfo {year} {2013})}\BibitemShut {NoStop}%
\bibitem [{\citenamefont {Zhang}\ \emph {et~al.}(2016)\citenamefont {Zhang},
  \citenamefont {Chen}, \citenamefont {Morrison}, \citenamefont
  {Vila-Comamala}, \citenamefont {Guizar-Sicairos},\ and\ \citenamefont
  {Robinson}}]{zhang2016phase}%
  \BibitemOpen
  \bibfield  {author} {\bibinfo {author} {\bibfnamefont {F.}~\bibnamefont
  {Zhang}}, \bibinfo {author} {\bibfnamefont {B.}~\bibnamefont {Chen}},
  \bibinfo {author} {\bibfnamefont {G.~R.}\ \bibnamefont {Morrison}}, \bibinfo
  {author} {\bibfnamefont {J.}~\bibnamefont {Vila-Comamala}}, \bibinfo {author}
  {\bibfnamefont {M.}~\bibnamefont {Guizar-Sicairos}}, \ and\ \bibinfo {author}
  {\bibfnamefont {I.~K.}\ \bibnamefont {Robinson}},\ }\href@noop {} {\bibfield
  {journal} {\bibinfo  {journal} {Nat. Comm.}\ }\textbf {\bibinfo {volume}
  {7}},\ \bibinfo {pages} {1} (\bibinfo {year} {2016})}\BibitemShut {NoStop}%
\bibitem [{\citenamefont {Wu}\ \emph {et~al.}(2021)\citenamefont {Wu},
  \citenamefont {Yoo}, \citenamefont {Suzana}, \citenamefont {Assefa},
  \citenamefont {Diao}, \citenamefont {Harder}, \citenamefont {Cha},\ and\
  \citenamefont {Robinson}}]{wu2021three}%
  \BibitemOpen
  \bibfield  {author} {\bibinfo {author} {\bibfnamefont {L.}~\bibnamefont
  {Wu}}, \bibinfo {author} {\bibfnamefont {S.}~\bibnamefont {Yoo}}, \bibinfo
  {author} {\bibfnamefont {A.~F.}\ \bibnamefont {Suzana}}, \bibinfo {author}
  {\bibfnamefont {T.~A.}\ \bibnamefont {Assefa}}, \bibinfo {author}
  {\bibfnamefont {J.}~\bibnamefont {Diao}}, \bibinfo {author} {\bibfnamefont
  {R.~J.}\ \bibnamefont {Harder}}, \bibinfo {author} {\bibfnamefont
  {W.}~\bibnamefont {Cha}}, \ and\ \bibinfo {author} {\bibfnamefont {I.~K.}\
  \bibnamefont {Robinson}},\ }\href@noop {} {\bibfield  {journal} {\bibinfo
  {journal} {npj Comput. Mater.}\ }\textbf {\bibinfo {volume} {7}},\ \bibinfo
  {pages} {1} (\bibinfo {year} {2021})}\BibitemShut {NoStop}%
\bibitem [{\citenamefont {Srivastava}\ \emph {et~al.}(2014)\citenamefont
  {Srivastava}, \citenamefont {Hinton}, \citenamefont {Krizhevsky},
  \citenamefont {Sutskever},\ and\ \citenamefont
  {Salakhutdinov}}]{srivastava2014dropout}%
  \BibitemOpen
  \bibfield  {author} {\bibinfo {author} {\bibfnamefont {N.}~\bibnamefont
  {Srivastava}}, \bibinfo {author} {\bibfnamefont {G.}~\bibnamefont {Hinton}},
  \bibinfo {author} {\bibfnamefont {A.}~\bibnamefont {Krizhevsky}}, \bibinfo
  {author} {\bibfnamefont {I.}~\bibnamefont {Sutskever}}, \ and\ \bibinfo
  {author} {\bibfnamefont {R.}~\bibnamefont {Salakhutdinov}},\ }\href@noop {}
  {\bibfield  {journal} {\bibinfo  {journal} {JMLR}\ }\textbf {\bibinfo
  {volume} {15}},\ \bibinfo {pages} {1929} (\bibinfo {year}
  {2014})}\BibitemShut {NoStop}%
\bibitem [{\citenamefont {Wilson}\ and\ \citenamefont
  {Izmailov}(2020)}]{wilson2020bayesian}%
  \BibitemOpen
  \bibfield  {author} {\bibinfo {author} {\bibfnamefont {A.~G.}\ \bibnamefont
  {Wilson}}\ and\ \bibinfo {author} {\bibfnamefont {P.}~\bibnamefont
  {Izmailov}},\ }\href@noop {} {\bibfield  {journal} {\bibinfo  {journal}
  {Advances in neural information processing systems}\ }\textbf {\bibinfo
  {volume} {33}},\ \bibinfo {pages} {4697} (\bibinfo {year}
  {2020})}\BibitemShut {NoStop}%
\bibitem [{\citenamefont {Jospin}\ \emph {et~al.}(2022)\citenamefont {Jospin},
  \citenamefont {Laga}, \citenamefont {Boussaid}, \citenamefont {Buntine},\
  and\ \citenamefont {Bennamoun}}]{jospin2022hands}%
  \BibitemOpen
  \bibfield  {author} {\bibinfo {author} {\bibfnamefont {L.~V.}\ \bibnamefont
  {Jospin}}, \bibinfo {author} {\bibfnamefont {H.}~\bibnamefont {Laga}},
  \bibinfo {author} {\bibfnamefont {F.}~\bibnamefont {Boussaid}}, \bibinfo
  {author} {\bibfnamefont {W.}~\bibnamefont {Buntine}}, \ and\ \bibinfo
  {author} {\bibfnamefont {M.}~\bibnamefont {Bennamoun}},\ }\href@noop {}
  {\bibfield  {journal} {\bibinfo  {journal} {IEEE Computational Intelligence
  Magazine}\ }\textbf {\bibinfo {volume} {17}},\ \bibinfo {pages} {29}
  (\bibinfo {year} {2022})}\BibitemShut {NoStop}%
\bibitem [{\citenamefont {Goodfellow}\ \emph {et~al.}(2015)\citenamefont
  {Goodfellow}, \citenamefont {Shlens},\ and\ \citenamefont
  {Szegedy}}]{goodfellow2014explaining}%
  \BibitemOpen
  \bibfield  {author} {\bibinfo {author} {\bibfnamefont {I.~J.}\ \bibnamefont
  {Goodfellow}}, \bibinfo {author} {\bibfnamefont {J.}~\bibnamefont {Shlens}},
  \ and\ \bibinfo {author} {\bibfnamefont {C.}~\bibnamefont {Szegedy}},\ }in\
  \href@noop {} {\emph {\bibinfo {booktitle} {International Conference on
  Learning Representations (ICLR)}}}\ (\bibinfo {year} {2015})\BibitemShut
  {NoStop}%
\bibitem [{\citenamefont {Weiss}\ \emph {et~al.}(2016)\citenamefont {Weiss},
  \citenamefont {Khoshgoftaar},\ and\ \citenamefont {Wang}}]{weiss2016survey}%
  \BibitemOpen
  \bibfield  {author} {\bibinfo {author} {\bibfnamefont {K.}~\bibnamefont
  {Weiss}}, \bibinfo {author} {\bibfnamefont {T.~M.}\ \bibnamefont
  {Khoshgoftaar}}, \ and\ \bibinfo {author} {\bibfnamefont {D.}~\bibnamefont
  {Wang}},\ }\href@noop {} {\bibfield  {journal} {\bibinfo  {journal} {J. Big
  Data}\ }\textbf {\bibinfo {volume} {3}},\ \bibinfo {pages} {1} (\bibinfo
  {year} {2016})}\BibitemShut {NoStop}%
\bibitem [{\citenamefont {Floridi}(2020)}]{floridi2020ai}%
  \BibitemOpen
  \bibfield  {author} {\bibinfo {author} {\bibfnamefont {L.}~\bibnamefont
  {Floridi}},\ }\href@noop {} {\bibfield  {journal} {\bibinfo  {journal}
  {Philos. Technol.}\ }\textbf {\bibinfo {volume} {33}},\ \bibinfo {pages} {1}
  (\bibinfo {year} {2020})}\BibitemShut {NoStop}%
\end{thebibliography}
\end{document}